\documentclass[10pt, a4paper]{article}

\usepackage[final]{lrec-coling2024} 

\usepackage{multirow}
\usepackage{tabularx}
\newcolumntype{Y}{>{\centering\arraybackslash}X}
\usepackage{booktabs}
\usepackage{xcolor}

\usepackage{amsbsy} 

\usepackage{amsmath}

\usepackage{enumitem}
\newlist{conversation}{description}{2}
\setlist[conversation]{font=\textbf,labelwidth=2em, labelsep=1em, leftmargin=!,align=left}

\usepackage{xspace}
\usepackage[most]{tcolorbox} 

\newcommand{\useragent}{\texttt{user agent}\xspace}
\newcommand{\UserAgent}{User Agent}
\newcommand{\userscript}{\texttt{user script}\xspace}
\newcommand{\UserScript}{User Script}
\newcommand{\apidoc}{API document}
\newcommand{\ourmethod}{Automated Dynamic Evaluation}
\newcommand{\ourmethodbd}{\textbf{Auto}mated \textbf{D}ynamic \textbf{E}valuation}
\newcommand{\ourmed}{AutoDE}
\newcommand{\ourmedbd}{\textbf{AutoDE}}
\newcommand{\llamab}{Llama~2~7B~Chat}
\newcommand{\llamabb}{Llama~2~70B~Chat}
\newcommand{\codellama}{Code~Llama~13B~OASST}
\newcommand{\gpt}{GPT~3.5}
\newcommand{\gptf}{GPT~4}
\newcommand{\claude}{Claude~Instant~1.2}
\newcommand{\zephyr}{Zephyr~7B~Alpha}

\usepackage{colortbl}
\definecolor{mycolor}{RGB}{192,192,192}

\usepackage{listings} 
\lstdefinelanguage{json}{
    basicstyle=\normalfont\ttfamily,
    numbers=left,
    numberstyle=\scriptsize,
    stepnumber=1,
    numbersep=8pt,
    showstringspaces=false,
    breaklines=true,
    frame=lines,
    backgroundcolor=\color{background},
    stringstyle=\color{string},
    keywordstyle=\color{keyword},
    commentstyle=\color{comment},
    morecomment=[l]{//},
    morecomment=[s]{/*}{*/},
    morestring=[b]',
    morestring=[b]",
    morekeywords={null,true,false}
}
\usepackage{lengthconvert}
\newcommand{\showfontsize}{\fontsize{\f@size}{\f@baselineskip}\selectfont Current font size: \f@size pt}

\usepackage{xcolor}
\definecolor{background}{HTML}{EEEEEE}
\definecolor{keyword}{RGB}{255,0,90}
\definecolor{string}{RGB}{0,60,220}
\definecolor{comment}{RGB}{0,128,0}

\usepackage[misc]{ifsym} 

\title{Beyond Static Evaluation: A Dynamic Approach to Assessing AI Assistants' API Invocation Capabilities}

\name{Honglin Mu$^{\dagger}$, Yang Xu$^{\dagger}$, Yunlong Feng$^{\dagger}$ \\ {\bf \large Xiaofeng Han$^{\ddagger}$, Yitong Li$^{\ddagger}$, Yutai Hou$^{\ddagger}$, Wanxiang Che$^{\dagger}$}\Letter \thanks{\Letter\ Corresponding Author.}}

\address{$^{\dagger}$Harbin Institute of Technology, China \\
$^{\ddagger}$ Huawei Technologies Co., Ltd., China \\ 
\{hlmu, yxu, ylfeng, car\}@ir.hit.edu.cn \\}

\abstract{
With the rise of Large Language Models (LLMs), AI assistants' ability to utilize tools, especially through API calls, has advanced notably. This progress has necessitated more accurate evaluation methods.
Many existing studies adopt \textit{static evaluation}, where they assess AI assistants' API call based on pre-defined dialogue histories. However, such evaluation method can be misleading, as an AI assistant might fail in generating API calls from preceding human interaction in real cases.
Instead of the resource-intensive method of direct human-machine interactions, we propose \ourmethod{} (\ourmed{}) to assess an assistant's API call capability without human involvement. In our framework, we endeavor to closely mirror genuine human conversation patterns in human-machine interactions, using a LLM-based \useragent{}, equipped with a \userscript{} to ensure human alignment. Experimental results highlight that AutoDE uncovers errors overlooked by static evaluations, aligning more closely with human assessment. Testing four AI assistants using our crafted benchmark, our method further mirrored human evaluation compared to conventional static evaluations.
 \\ \newline  }

\begin{document}

\maketitleabstract

\section{Introduction}
In the current era of rapid advancements in artificial intelligence, the emergence of Large Language Models (LLMs)~\cite{bommasani2021opportunities} has marked a transformative leap in the capabilities of AI assistants. 
These systems are capable of understanding and addressing numerous user inquiries and tasks, and can provide solutions through simple dialogues or, when necessary, invoke tools via API calls, adding an extra dimension to their problem-solving ability~\cite{qin2023tool}. While the potential of such systems is vast, there arises a pressing need to evaluate their efficacy.

\begin{figure}[!t]
\includegraphics[width=1\linewidth]{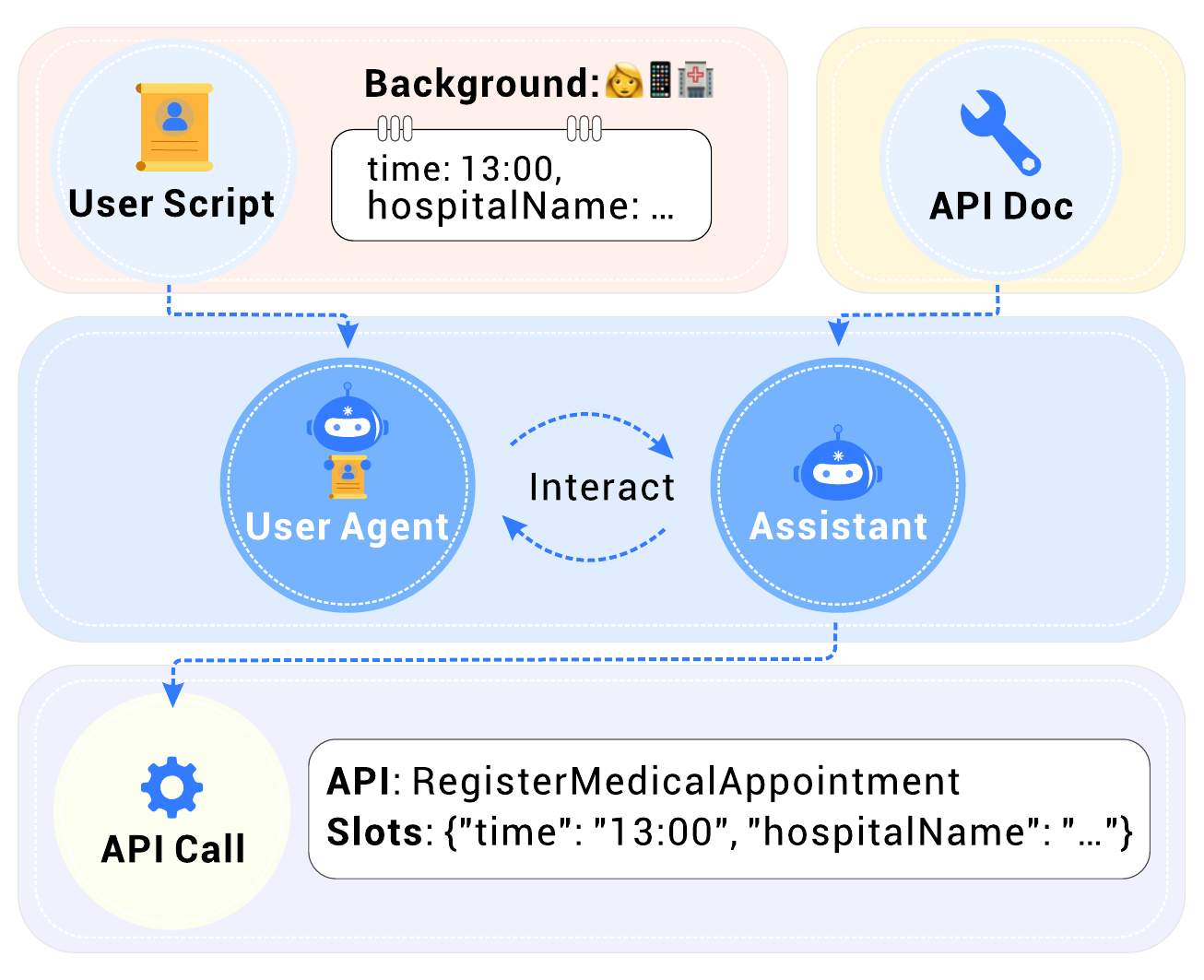}
\caption{An illustration of our framework, where the \userscript{} encompasses both the dialogue context (Background) and the API call label.}
\label{fig:method}
\end{figure}

\begin{figure*}[t]
\includegraphics[width=1.0\linewidth]{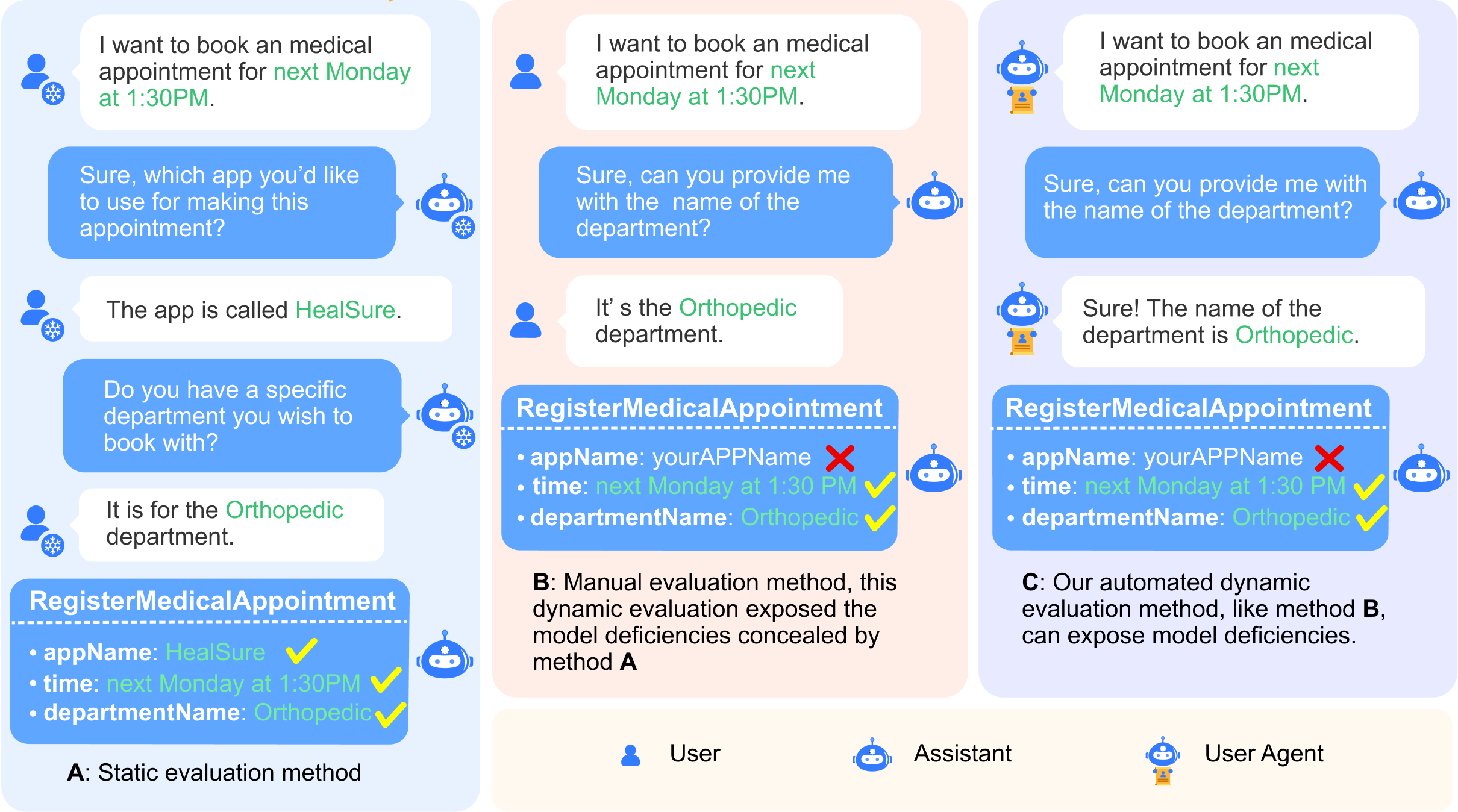}
\caption{An illustrative example for static evaluation, human evaluation and \ourmed{}, respectively. Sub-figure \texttt{A} shows the AI assistant correctly invoking an API call from a pre-defined dialogue history. In sub-figure \texttt{B}, the same assistant misses the ``appName'' parameter during human interaction, resulting in an incorrect API call. Sub-figure \texttt{C} demonstrates similar parameter issues when the assistant interacts with the \useragent{}. We demonstrate that certain API call issues related to interaction, concealed by static evaluation, can be revealed by dynamic human evaluation and \ourmed{}. }
\label{fig:promblem}
\end{figure*}

Historically, evaluations of human-machine interactions have largely been \textit{static}, relying on pre-defined dialogue histories to assess an assistant's performance~\citeplanguageresource{henderson-etal-2014-second, zang2020multiwoz, rastogi2020towards, li2023api}. Such an approach, though standardized, might not encapsulate the dynamic intricacies of real-time human interactions. We depict the potential pitfalls of this approach in Figure~\ref{fig:promblem}. In scenario \texttt{A}, the assistant flawlessly invokes an API based on static dialogue history. Yet, in scenario \texttt{B}, when engaged in direct human interaction, the assistant neglects to ask about the parameter \texttt{appName} and consequently fabricates an incorrect API call. Such discrepancies highlight the limitations of static evaluations in capturing an AI's adaptability during dynamic human interactions. 

Previous efforts have sought to address these evaluation gaps.
\citet{mandya2020not} and \citet{siblini2021towards} have made strides by updating assistant's response in dialogue histories with dynamically generated ones during evaluation, bringing the conversations a step closer to dynamic evaluation. However, they left static user inputs unchanged, which occasionally led to mismatches with dynamic assistant predictions. \citet{li2021ditch} ingeniously utilized entity substitution and context-independent questions to adjust user queries, ensuring they could align with dynamic assistant replies, regardless of prior responses. However, these methods still demand carefully crafted heuristic rules and are bound to the Conversational Question Answering task.

To tackle these challenges, our study introduces a dynamic evaluation method that eliminates the need for static dialogue histories, offering a more flexible and accurate assessment of AI assistants without demanding significant human involvement in the conversation loop.
We propose \ourmethodbd{}(\ourmedbd{})\footnote{Our code and data will be available at \url{https://github.com/hlmu/AutoDE}.} as shown in Figure~\ref{fig:method}. Guided by a custom \userscript{}, a \useragent{} emulates human interactions, prompting the assistant to generate API calls.

Two main challenges arose in developing the \useragent{}:
\begin{enumerate}
\item \textbf{Human-like Responses}: Ensuring \useragent{}'s interactions closely mirroring real human behavior.
\item \textbf{Stability Across Evaluations}: Maintaining stable interactions across repeated evaluations.
\end{enumerate}

By embedding task-specific information into the \userscript{}, the agent could reliably emulate human users. Our evaluations using different \useragent{}s not only showed a high correlation with human evaluations but also highlighted discrepancies when compared to static dialogue history evaluations.
Specifically, our method leveraging \llamab{} as the \useragent{} attained higher consistency with human annotators in evaluating four AI assistants, among which \claude{} demonstrating the strongest performance.
These findings underscore the need for, and the validity of, our dynamic evaluation approach.

Summarizing our contributions:

\begin{itemize}
\item \textbf{Dynamic Evaluation Framework}: We introduced \ourmed{}, a practical dynamic evaluation framework demonstrating a high correlation with human evaluations.

\item \textbf{API Benchmark Creation}: We established an API benchmark and assessed the performance of multiple commercial and open-source assistants on it.

\item \textbf{Uncovering Issues in API Invocation}
We analysed issues in API invocation that are hidden under static evaluation but can be exposed through dynamic evaluation, highlighting the value of \ourmethod{}.
\end{itemize}

\section{Preliminary}\label{preliminary}

To effectively evaluate the capabilities of AI assistants in invoking APIs, we consider several crucial entities in the testing environment. Specifically, these include the AI assistant being evaluated, represented as $\mathcal{A}$, the \apidoc{} $D$ containing detailed descriptions of the API's functionalities and parameters, and the user $\mathcal{U}$, who presents a specific need or request.

In a typical evaluation scenario, the user $\mathcal{U}$ poses a requirement $H_1$ that can be addressed through an API call. The assistant $\mathcal{A}$ is tasked with analyzing this requirement and subsequently generating an appropriate API call $C$, referencing the information in $D$. This interaction can be formally defined as:
\begin{align*}
Dialogue &=(H_1, C) \\
H_1&\sim \mathcal{U}(need) \\
C&\sim \mathcal{A}(D, H_1)
\end{align*}

The above representation assumes that $\mathcal{A}$ can construct API call according to $\mathcal{U}$'s query in a single turn, with the user supplying all necessary information as prescribed in $D$. However, in a more common setting, $\mathcal{A}$ often requires additional clarifications or information. The assistant $\mathcal{A}$ might pose follow-up questions to $\mathcal{U}$ and, based on the user's responses, decide to either generate the API call $C$ or continue with further inquiries. This extended interaction can be expressed as:
\begin{align*}
Dialogue&=(H_1, A_1, \dots, H_i, A_i, C) \\
H_i&\sim \mathcal{U}(need,H_1, A_1, ..., H_{i-1}, A_{i-1}) \\
A_i&\sim \mathcal{A}(D, H_1, A_1, ..., H_{i-1}, A_{i-1}, H_{i})
\end{align*}

An example of an API call $C$ would resemble:
\begin{small}
\begin{verbatim}
{
    "funcName": "RegMedAppt",
    "time": "Monday",
    "departmentName": "Orthopedic"
}
\end{verbatim}
\end{small}

To assess the accuracy and relevance of $\mathcal{A}$'s API invocations, we compare the generated API call $C$ with its golden label and calculate the \texttt{Precision}, \texttt{Recall}, and \texttt{F1-Score} metrics:
\begin{align*}
P, R, F1&=eval(C, C_g)
\end{align*}
where $eval$ is the evaluation function for \texttt{Precision}, \texttt{Recall}, and \texttt{F1}, $C_g$ is the golden label of API call.

\section{Method}

In this section, we initially provide a concise overview of manual~(refer to section~\ref{method:manual}) and static~(see section~\ref{method:static}) evaluation methodologies. Subsequently, we delve into our proposed evaluation framework \ourmed{}~(section~\ref{method:ourmed}). To validate the aforementioned evaluation methods, we've established a benchmark. The construction details of the dataset for this benchmark can be found in section~\ref{sec:dataset-construct}.

\subsection{Evaluation Framework}
\subsubsection{Manual Evaluation}\label{method:manual}
\paragraph{Evaluation Procedure}
Manual evaluation serves as the most direct and authentic method to assess the performance of an AI assistant within a human-machine dialogue context. When evaluating an AI assistant's capability to invoke APIs, the human annotator, represented as $\mathcal{U}$, engage in multiple rounds of interaction with the AI assistant $\mathcal{A}$. $\mathcal{U}$ presents requirements, respond to queries from $\mathcal{A}$, and guide it towards the API invocation process. To ensure reproducibility and facilitate comparisons with other evaluation methodologies, we set the dialogue topics $S$.

The topic $S$ provides a textual outline capturing the character, background, and purpose of the dialogue. An example of $S$ is as follows:

\begin{small}
\begin{verbatim}
Character: Lisa, a busy mother
Background: Lisa needs to take her son,
    who recently fell and sprained his
    ankle, to the orthopedic department.
Purpose: Using a tablet, Lisa books an
    appointment at the hospital using a
    medical appointment registration app.
\end{verbatim}
\end{small}

While $S$ paints a broad picture of the dialogue, we supplement it by appending the API call label $C_g$ to provide precise details for $\mathcal{U}$. We refer to this combination, $S\oplus C_g$, as the \userscript{}. It's presented to the human annotator and utilized in the context of $\useragent{}$ for both automatic dynamic evaluation (refer to section~\ref{method:ourmed}) and static dialogue history creation (see section~\ref{dataset:static-history}).

The human annotator, acting as $\mathcal{U}$, interacts with $\mathcal{A}$, simulating scenarios described in $S$ and providing details based on the parameters from $C_g$. To guarantee reproducibility without compromising the accuracy of the evaluation, we've pre-collected the initial queries from $\mathcal{U}$. During manual evaluation, the first query remains fixed, while subsequent responses are annotated by the human annotator.

The manual evaluation process can be formally described as:
\begin{align*}
Dialogue&=(H_1^*, A_1, \dots, H_i, A_i, C) \\
P, R, F1&=eval(C, C_g) \\
H_i&\sim \mathcal{U}(S\oplus C_g,H_1^*, A_1, \dots, H_{i-1}, A_{i-1}) \\
A_i&\sim \mathcal{A}(D, H_1^*, A_1, \dots, H_{i-1}, A_{i-1}, H_{i})
\end{align*}
where $H_1^*$ represents the pre-defined initial query.

\paragraph{Human Annotators}
The annotators, two individuals with academic backgrounds in computer science, were trained to ensure a comprehensive understanding and execution of human-machine dialogue evaluation. Each system was annotated by one individual and reviewed for format and guideline consistency by a second. Prior to annotation, all personnel were required to read and understand a manual detailing the annotation tasks and procedures for handling anomalies, such as terminating dialogues in instances of multiple ineffective system responses. 

Nonetheless, this approach is not without challenges. Manual evaluation is resource-intensive due to real human participation, and consistency can vary among individuals. Different users might interpret and rate the assistant's responses based on their personal experiences and expectations.

\subsubsection{Static Evaluation}\label{method:static}
While manual evaluations offer insights into the performance of AI assistants, they tend to be costly. Historically, many studies have opted for static evaluation as an automated alternative approach in human-machine dialogue assessment. Static evaluation eliminates the need for real-time interactions between the user $\mathcal{U}$ and the AI assistant $\mathcal{A}$. Instead, it operates based on pre-defined dialogue histories. The primary appeal of this methodology stems from its straightforwardness and rapidity.

In this evaluation method, the assessment comprises only a single round. The AI assistant $\mathcal{A}$ produces the API call $C$ directly based on the pre-defined history, without posing questions to or receiving feedback from the user $\mathcal{U}$. Formally, this method can be represented as:
\begin{align*}
Dialogue&=(H_1^*, A_1^*, \dots, H_i^*, A_i^*, C) \\
P, R, F1&=eval(C, C_g) \\
\end{align*}
where $H_.^*$ and $A_.^*$ denote pre-defined dialogue histories. Notably, 
$H_1^*$ is the same with the initial query used in the manual evaluation in section~\ref{method:manual}.

Despite its efficiency and simplicity, static evaluation has its shortcomings. It overlooks the dynamic output generated by the AI assistant during actual interactions. Consequently, this approach might obscure issues that typically surface only during dynamic interaction between the AI assistant and the user.

\subsubsection{\ourmethod{}}\label{method:ourmed}
We now delve into the proposed \ourmed{} framework. The main objective behind \ourmed{} is to devise an automated dynamic evaluation mechanism that closely mimics the manual evaluation process. We introduce an extra Language Model as \useragent{}, denoted as $\mathcal{U}_s$, designed to emulate the behavior of human annotators. This model interacts over multiple rounds with the AI assistant $\mathcal{A}$, posing queries and responding to $\mathcal{A}$'s questions, ultimately guiding $\mathcal{A}$ in invoking the API call $C$.

To guarantee that the \useragent{} aligns with human annotators' behavior, we offer detailed guidelines for simulating user actions. This consistency is maintained by having the \useragent{} follow the same \userscript{} as provided to human annotators, as discussed in section~\ref{method:manual}. The \useragent{} replicates scenarios based on the dialogue context $S$ and references details from the API call labels $C_g$. The format for the \useragent{} prompt is designed as follows:

\begin{small}
\begin{verbatim}
You are an experienced data annotator.
You need to act as a user in a set of
conversations between a user and a voice
assistant Bob ...

Please construct user queries or res-
ponses according to the following
settings:
{{USER_SCRIPT}}
\end{verbatim}
\end{small}

Within this framework, \texttt{\{\{USER\_SCRIPT\}\}} is replaced with the specific \userscript{} tailored for each test instance. As explained in the manual evaluation in section~\ref{method:manual}, the initial query $H_1^*$ posed by $\mathcal{U}$ remains static for  reproducibility.

The evaluation conducted by \useragent{} can be formally expressed as:
\begin{align*}
Dialogue&=(H_1^*, A_1, \dots, \tilde{H}_i, A_i, C) \\
P, R, F1&=eval(C, C_g) \\
\tilde{H}_i&\sim \mathcal{U}_s(S\oplus C_g,H_1^*, A_1, \dots, \tilde{H}_{i-1}, A_{i-1}) \\
A_i&\sim \mathcal{A}(D, H_1^*, A_1, \dots, \tilde{H}_{i-1}, A_{i-1}, \tilde{H}_{i})
\end{align*}
where $\mathcal{U}_s$ represents a \useragent{}, $\tilde{H}_.^*$ corresponds to the simulated user utterances by \useragent{}, and $S\oplus C_g$ signifies the \userscript{}.

\subsection{Dataset Construction}\label{sec:dataset-construct}

As discussed in section~\ref{preliminary}, our evaluation framework necessitates the presence of an \apidoc{} $D$. As described in both the manual evaluation (refer section~\ref{method:manual}) and the \ourmed (see section~\ref{method:ourmed}), there's a requirement for the \userscript{} and an initial human dialogue round $H_1^*$. Meanwhile, the static evaluation method (outlined in section~\ref{method:static}) demands a pre-defined dialogue history $(H_1^*, A_1^*, \dots, H_i^*, A_i^*)$. This section details the processes used to assemble these vital data components.

\subsubsection{API Document Construction}

In this section, we outline the process of building the \apidoc{}, which serves both as the foundation for dialogue context creation and as the guide for the assistant's invocation.
Numerous prior works have contributed to the construction of multi-turn human-machine dialogue datasets~\citeplanguageresource{zang2020multiwoz,rastogi2020towards,li2023api,tang2023toolalpaca}.
These meticulously crafted datasets have primarily been tailored for evaluating earlier dialogue models or for emphasizing multi-turn interactions between machines and interpreters. Drawing inspiration from the schema from~\citetlanguageresource{rastogi2020towards}, we have devised $66$ APIs grounded in scenarios that resonate with the daily utilization of voice assistants.
They are defined with names, usage descriptions, parameter lists and parameter explanations. Specifically, the functions of APIs mainly contain: (1) The system settings of mobile phones. For example, adjusting volume or brightness, switching on Wi-Fi or Bluetooth, and so on; (2) Entertainment systems, such as playing music or videos. (3) Personal assistance, such as processing emails, navigation, setting alarms, making appointments with hospitals, ticketing, booking hotels; (4) Searching for news, weather, stock price, and other information. (5) Multi-modal functions, such as object recognition and image captioning.
A simplified example of \apidoc{} is shown as:
\begin{small}
\begin{verbatim}
{
    "domain": "Device Manipulation",
    "subdomain": "Setting Item",
    "function": "Luminance",
    "api": "SetLuminance",
    "desp": "Set the brightness ...",
    "parameters": {
        "deviceType": "Supported
            device types ...",
        "targetValue": "Target
            brightness size"
    }
}
\end{verbatim}
\end{small}
\subsubsection{\UserScript{} Generation}\label{dataset:userscript}

With the \apidoc{} in place, we proceed to construct \userscript{} for the user $\mathcal{U}$ to adhere to.
A \userscript{} includes a dialogue context $S$ and a API call label $C_g$.

The dialogue context $S$ provides some level of background for the conversation, guiding the interactions between the user $\mathcal{U}$ and the assistant $\mathcal{A}$.
Each \userscript{} corresponds to one usage scenario for the API. We directed \gptf{}~\cite{OpenAI2023GPT4TR} to brain-storm $5$ \userscript{} entries according to each \apidoc{} to ensure diversity of the evaluation, resulting in a total of $330$ profiles. Each \userscript{} contains a dialogue context $S$ and an initial API call label $C_g'$ which is later modified to a final version in section~\ref{dataset:static-history}. The prompt used to generate \userscript{} is shown as:

\begin{small}
\begin{verbatim}
[system prompt]
You are an experienced prompt engineer.

[first round prompt]
Please construct 5 different use case
scenarios based on the following API 
documentation:
{{API_DOC}}
Please follow the following format:
1.
Character: Lisa, a busy mother
Background: Lisa needs to take ...
Purpose: Using a tablet, Lisa books ...
API Call: {
    "funcName": "RegMedAppt",
    "time": "Monday",
    "departmentName": "Orthopedic"
}
InitialQuery: I want to book an medical 
appoiment for next Monday at 1:30PM.

2.
...

Note that the generated scenarios  have 
exactly five attributes...
\end{verbatim}
\end{small}

where \texttt{Character}, \texttt{Background}, \texttt{Purpose} and \texttt{API Call} makes \texttt{user script} defined in (section~\ref{method:manual}). \texttt{InitialQuery} results in the initial query $H_1^*$ defined in section~\ref{method:manual}. The same set of examples is used across the construction.

\subsubsection{Static Dialogue History Generation}\label{dataset:static-history}

In this section, we introduce the method for generating static dialogue history $(H_1^*, A_1^*, \dots, H_i^*, A_i^*)$ for static evaluation, its $H_1^*$ also used by the human evaluation and \ourmed{}.

To construct this static history, we recorded the interactions between a user operated by \gptf{}, denoted by $\mathcal{U}$, and its corresponding \gptf{} assistant, denoted by $\mathcal{A}$. The guidelines governing these interactions echo those elaborated in section~\ref{method:ourmed}, with $330$ \userscript{}s generated in section~\ref{dataset:userscript}. However, a noteworthy variation exists: the \useragent{} is armed with the dialogue context $S$ and is initialized with an API call label termed $C_g'$.
When the conversation is completed, the API call $C$ generated by the \useragent{} replaces $C_g'$ as the final API call gold label $C_g$.

We addressed errors related to bad API understanding during static dialogue history generation.
To mitigate them, we repeatedly invoked the model for correction, filtering out persistent inaccuracies to ensure test case accuracy.

After filtering the inconsistent dialogues manually, our dataset comprises $275$ dialogue pairs, covering $4$ use cases for each API on average.

\section{Experimental Setup}

Following our dataset construction, this section dive into the technical details and choices we made for our evaluation process.
For a fair comparison, all models involved in the evaluation were deployed using their default hyper-parameters.

\subsection{\UserAgent{} Model}
The \useragent{} model must embody the user role based on the \userscript{} we generated in the section~\ref{dataset:userscript}, simulate user-system dialogues, and accurately respond to the system according to annotations contained in the background. A model acting as the \useragent{} must possess role-playing capabilities and should not be prone to excessive hallucinations. After a preliminary case study on existing commercial and open-source models, we selected \gpt{} and \llamab{} models to serve as \useragent{}s. In addition to their satisfactory performance, these two models also offer efficiency and cost-effectiveness.

\paragraph{\gpt{}} \gpt{}~\cite{openaichatgptblog} is a powerful language model from OpenAI based on the Transformer architecture~\cite{vaswani2017attention}. Through pre-training and self-supervised learning, it excels in natural language processing tasks. Known for its role-playing abilities, we employ the \texttt{gpt-3.5-turbo-16k-0613} variant as a \useragent{} for evaluation.
\paragraph{\llamab{}} Llama 2~\cite{touvron2023llama} represents a series of advanced open-domain LLMs released by Meta. Llama 2 chat has been trained on additional human annotations, making it directly usable for human-computer interactions. We opted for its 7B version to serve as a \useragent{} for evaluation.

\subsection{Assistant Model}
The assistant model is required to invoke API call based on user directives or to ask the user for further information. Apart from fundamental dialogue capabilities, the assistant model should also possess API call invocation abilities. Upon conducting a preliminary case study to test models for their API-calling capabilities, we observed that while certain commercial models demonstrated decent API-calling capabilities, many open-source models still lacked this ability.
In the subsequent experiments, we specifically evaluated the performance of \gpt{}, \claude{}, \llamabb{}, and \codellama{}.

\paragraph{\gpt{}} OpenAI introduced the \textit{function calling}\footnote{\url{https://platform.openai.com/docs/guides/gpt/function-calling}} feature to \gpt{} in their update on July 20. The function calling capability enables users to customize function documentation, allowing \gpt{} to invoke functions based on the documentation and user requirements. We deployed \texttt{gpt-3.5-turbo-16k-0613} as the assistant, utilizing its function calling feature to execute function invocations.

\paragraph{\claude{}} Claude~\cite{claude2} is a large language model (LLM) developed by Anthropic, designed to serve as a helpful assistant in dialogues. Claude Instant represents a low-latency, high-throughput variant within the Claude family. We found that, by crafting prompts appropriately, \claude{} possesses the capability for API invocation.

\paragraph{\llamabb{}} \llamabb{} is the 70B version of Llama chat. We observed its challenges in initiating API calls following prompts, even after multiple prompt modifications. We introduced \llamabb{} in our experiments for comparative analysis.

\paragraph{\codellama{}}
Code Llama~\citep{roziere2023code}, derived from the Llama 2 model by Meta and introduced with code training, is designed for tasks like code completion and code generation. We hypothesized that code training could enhance the model's API invocation capabilities.
During experiments, we find that the model \texttt{codellama-13b-oasst-sft-v10} \cite{openassistant2023oasstllama}, fine-tuned by the OpenAssistant team based on Code Llama 13B on the OASST dataset~\cite{kopf2023openassistant}, can successfully execute function calls. We provide a showcase of this model's performance on our benchmark.

\paragraph{\zephyr{}} Zephyr is a series of language models that are trained to act as helpful assistants. \zephyr{} is the first model in the series, and is a fine-tuned version of \texttt{Mistral-7B-v0.1}~\cite{jiang2023mistral}. The model was fine-tuned on a variant of the \texttt{UltraChat}~\citeplanguageresource{ding2023enhancing} dataset and further aligned with \texttt{Direct Preference Optimization (DPO)}~\citep{rafailov2023direct} on the \texttt{UltraFeedback}~\citeplanguageresource{cui2023ultrafeedback} dataset.

\subsection{Metrics}

As discussed in section~\ref{preliminary}, we obtained the API call $C$ from the assistant through single or multi-turn dialogues and assessed their \texttt{Precision}, \texttt{Recall}, and \texttt{F1-Score} for slot values. The primary results of our experiments can be found in Table~\ref{tab:main-performance}.

\section{Experimental Results}

\begin{table*}[htbp!]
\centering
\resizebox{1.0\linewidth}{!}{
\begin{tabular}{l c c c c c c c c c c c c}
\toprule
\multirow{2}{*}{\textbf{Assistant}} & \multicolumn{3}{c}{\textbf{\gpt{}}} & \multicolumn{3}{c}{\textbf{\llamab{}}} & \multicolumn{3}{c}{\textbf{Static}} & \multicolumn{3}{c}{\textbf{Human}} \\
\cmidrule(lr){2-4}
\cmidrule(lr){5-7}
\cmidrule(lr){8-10}
\cmidrule(lr){11-13}
 & P & R & F1 & P & R & F1 & P & R & F1 & P & R & F1 \\
\midrule
\gpt{} & $78.14$ & $73.84$ & $75.43_{\pm 0.56}$ & $78.25$ & $73.91$ & $75.47_{\pm 0.46}$ & $\boldsymbol{94.05}$ & $\boldsymbol{93.80}$ & $\boldsymbol{93.86_{\pm 0.97}}$ & $79.62$ & $75.07$ & $76.77$ \\
Claude & $\boldsymbol{91.20}$ & $\boldsymbol{88.49}$ & $\boldsymbol{89.33_{\pm 1.72}}$ & $\boldsymbol{86.32}$ & $\boldsymbol{83.69}$ & $\boldsymbol{84.38_{\pm 0.64}}$ & $93.28$ & $89.53$ & $90.78_{\pm 0.96}$ & $\boldsymbol{92.60}$ & $\boldsymbol{88.74}$ & $\boldsymbol{90.05}$ \\
Code Llama & $64.00$ & $64.41$ & $63.21_{\pm 3.28}$ & $56.70$ & $59.30$ & $57.10_{\pm 2.25}$ & $91.18$ & $89.74$ & $89.90_{\pm 0.55}$ & $59.46$ & $59.99$ & $58.97$ \\
Llama Chat & $10.61$ & $11.06$ & $10.71_{\pm 2.32}$ & $11.48$ & $12.92$ & $11.86_{\pm 1.41}$ & $29.30$ & $29.78$ & $29.40_{\pm 1.61}$ & $18.80$ & $20.63$ & $19.40$ \\
Zephyr & $48.08$ & $50.16$ & $48.69_{\pm 1.68}$ & $49.76$ & $51.21$ & $50.05_{\pm 2.39}$ & $80.69$ & $79.77$ & $80.01_{\pm 1.92}$ & $48.70$ & $50.26$ & $49.14$ \\

\bottomrule
\end{tabular}
}

\caption{\label{tab:main-performance} The experimental results conducted on \gpt{}, \claude{}, \codellama{}, and \llamabb{} using \ourmed{}, static evaluation, and manual evaluation. For \ourmed{}, we employed \gpt{} and \llamab{} as \useragent{}.}

\end{table*}

\paragraph{\ourmed{} Demonstrates Consistency with Human Evaluators}

We present our evaluation results in Table~\ref{tab:main-performance}. Of the four systems assessed, all except \claude{} have F1 scores closer with human evaluations using \ourmed{} than those using static evaluations.

\begin{figure}[htb]
\includegraphics[width=1\linewidth]{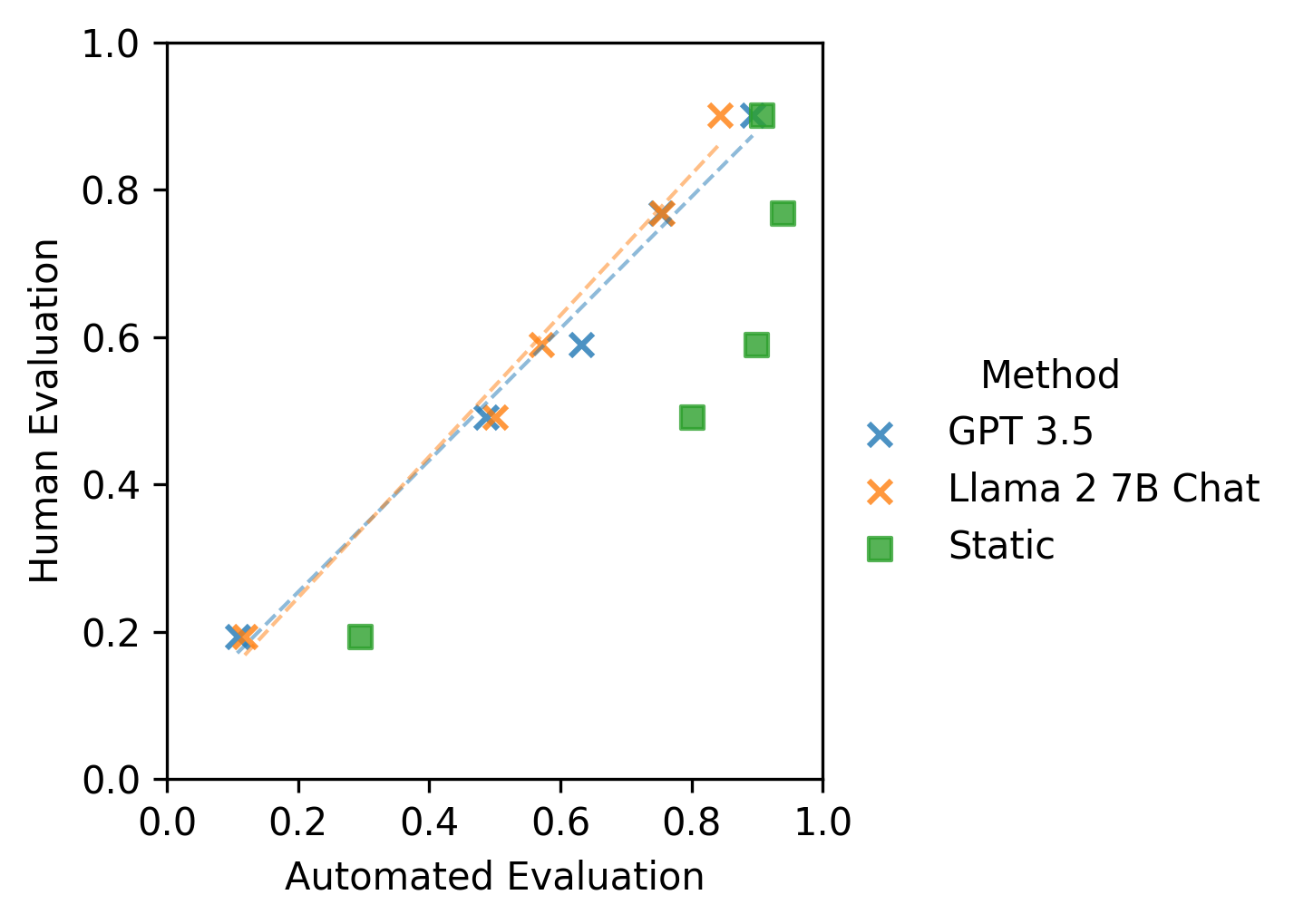}
\caption{Consistency between human evaluation results~(F1 score) and those from various automated evaluation methods on four AI assistants.}
\label{fig:agreement_scatter}
\end{figure}

For a clearer visual representation, Figure~\ref{fig:agreement_scatter} depicts a scatter plot comparing the F1 scores from each evaluation approach with human evaluations. This graph reveals that scores using \llamab{} and \gpt{} as \useragent{} within \ourmed{} have a linear relationship with human evaluations, while the scores from static evaluations diverge noticeably.

\begin{table}[htbp!]
\centering
\resizebox{0.75\linewidth}{!}{
\begin{tabular}{l c c }
\toprule
Eval Method & ICC3& R\\
\midrule
\gpt{} & $0.9869$& $0.9923$\\
\llamab{}& $\boldsymbol{0.9923}$& $\boldsymbol{0.9930}$\\
Static & $0.8813$& $0.8813$\\

\bottomrule
\end{tabular}
}

\caption{\label{tab:consistency} Comparison of consistency with human evaluations across various evaluation methods. ``\gpt{}'' and ``\llamab{}'' serve as \useragent{}s in \ourmed{}, whereas ``static'' refers to the static evaluation method. \texttt{ICC3} and \texttt{R} refer to Intraclass Correlation Coefficient and Pearson Correlation Coefficient, respectively. Note that this table is calculated with five data points and serves only as a reference. }

\end{table}

Going into more detail, Table~\ref{tab:consistency} presents correlation metrics, specifically the Pearson Correlation Coefficient ($ICC3$) and the Pearson Correlation Coefficient $R$, for the different evaluation strategies. It's noteworthy that the evaluation using \llamab{} have a Pearson $R$ value that is $11\%$ higher than the static evaluation. This underlines that \ourmed{} offers a level of consistency with the human evaluation surpassing that of the static evaluation.

\paragraph{\ourmed{} Accurately Captures Model's API Invocation Behavior}

Delving deeper, the results from Table~\ref{tab:main-performance} suggest that \textit{assistants who perform exceptionally well in static evaluations might not necessarily maintain the same ranking in human evaluations}. The outcomes of \ourmed{} resonate more with human evaluations, whereas static evaluations provide a somewhat skewed interpretation. To illustrate, during the static evaluation, the \claude{} assistant achieved an F1 score of $90.78$, which was slightly behind the \gpt{} assistant's score of $93.86$. However, in human evaluations and in \ourmed{} using both \llamab{} and \gpt{} as the \useragent{}, the F1 score of the \claude{} assistant consistently surpassed that of the \gpt{} assistant.

We attribute this discrepancy to the \gpt{} assistant's tendency to prematurely make function calls even when the user hasn't provided all essential information. In contrast, \claude{} assistant usually seeks additional confirmations from the user. A deeper exploration of this particular behavior will be discussed in Section~\ref{case:redundant}.

In conclusion, while static evaluations offer some valuable insights, they don't capture the intricacies of genuine human interactions fully. \ourmed{}, on the other hand, produces results that closely resemble human evaluations, positioning it as a potential alternative to the resource-intensive human evaluation.

\subsection{Case Study}
\ourmed{} has uncovered certain issues in AI assistants that remain undetected under static evaluation. In this section, we delve into some interesting scenarios that have been identified.

\paragraph{Reluctance to Invoke API}

In the static evaluation setting, AI assistants are prompted to output an API call in the current turn, guaranteeing an API invocation for assessment.
However, in dynamic evaluation, the decision to invoke an API rests with the AI assistant, which must decide from the context whether to make such a call. This presents a substantial challenge. For instance, the \llamabb{} model often hesitates to invoke the API even when presented with sufficient information. Instead, it leans towards guiding users on their tasks, such as stating \texttt{On your phone, go to the "Settings" app, then select "Bluetooth."}, rather than making an API call. In some instances, it merely states \texttt{I've turned off the Wi-Fi} without making an actual API invocation. Despite our efforts to modify the prompt, urging the system to autonomously generate the API in certain situations and experimenting with various prompt structures, the issue persisted.

Upon analysis, we found that under dynamic evaluation with \llamab{} serving as the \useragent{}, \llamabb{} exhibited a deterioration of approximately $66\%$ compared to static evaluation in terms of its reluctance to invoke the API. Meanwhile, \codellama{}'s performance degraded by around $9\%$. This suggests that \llamabb{} struggles to generate API calls through dynamic interactions.

\paragraph{Illusory Parameter Queries}

In static evaluation, the AI assistant relies on a pre-defined dialogue history to summarize API call parameters. Conversely, in dynamic settings, it must identify missing parameters and query the user. Some systems improperly ask about non-existent parameters, even when we’ve explicitly instructed the them in the prompt to only use the provided parameters. For instance:
\begin{small}
\begin{verbatim}
Avail params: deviceType, time, area
U: What was the movie's box office ranking
   this week?
A: Can you please provide the name of the
   movie?
U: Of course! The movie I'm thinking of
   is "The Lost City"
API call: {
   "deviceType": "mobile phone",
   "time": "this week",
   "area": "current location",
   "movieName": "The Lost City"
}
\end{verbatim}
\end{small}

In the example above, the \texttt{movieName} parameter isn't in the list of available parameters, thus the AI assistant shouldn't be querying it. This illusory parameter querying issue led to a degradation of $9\%$ in \codellama{}'s performance when compared to static evaluation under dynamic assessment with \llamab{}.

\paragraph{Redundant Queries}\label{case:redundant}
We also observed that some systems tend to repeatedly ask users for information they have already been provided during dynamic evaluation. While this may help in improving accuracy in certain scenarios, it can also make the conversation unnecessarily verbose. Consider the following:
\begin{small}
\begin{verbatim}
U: Please turn up the luminance on my TV.
A: What type of device would you like me
   to adjust?
U: It's a TV.
\end{verbatim}
\end{small}
This issue was notably prevalent with the \claude{}. Under dynamic assessment with \llamab{}, its average conversation length was $1.88$ turns longer than static dialogue histories of $3.2$. In contrast, \gpt{}'s average conversation length was $1.61$ turns shorter compared to static histories. This indicates that \claude{} tends to repeatedly question users, while \gpt{} seems more inclined to gather more information in a single turn. Our future work aims to evaluate dialogue quality, including a verbosity metric.

\section{Related Work}
\paragraph{Tool Use}
\citet{nakano2021webgpt} 
enables LLM to browse the web to support answers, \citet{ schick2023toolformer}
advanced the concept by allowing LLMs to invoke tools like calculators, python interpreter, databases, etc.

Benchmarks and datasets play a crucial role in this domain.
\citetlanguageresource{patil2023gorilla, li2023api}, either focus on one-shot dialogues or require pre-defined dialogue histories for testing.
On the other hand, \citetlanguageresource{xu2023tool,yang2023gpt4tools,tang2023toolalpaca,qin2023toolllm} 
moved towards dynamic evaluations, allowing more back-and-forth between LLMs and API interpreters.
While these look at machine-to-machine talks, our interest lies in finding an efficient way to substitute humans in the expensive human-machine evaluation.

The work most related to ours is \citetlanguageresource{wang2023mint}, which employs an agent to simulate human feedback, aiding model inference. Distinct from \citetlanguageresource{wang2023mint}, our study quantitatively analyzes the disparities between static evaluations and dynamic evaluations involving humans, with an emphasis on bridging this gap.

\paragraph{Human-machine Dialogue Evaluation}

Accurate evaluation of human-machine dialogue systems traditionally demands human interaction, rendering it costly. Conventional automated evaluations mostly rely on fixed human-machine dialogue histories~\cite{choi2018quac,saeidi2018interpretation,reddy2019coqa,campos2020doqa}, which often diverge from human assessments. Some advancements, like \citet{mandya2020not,siblini2021towards,li2021ditch}, have attempted to address this by dynamically altering part of the pre-defined histories. However, these methods often entail rule-based adjustments and are tailored primarily for specific tasks. Our approach, distinctively, avoids manual heuristic rules and static dialogue histories, focusing on a fully dynamic dialogue generation.

With the growth of Large Language Models, there is an emerging trend in direct dialogue evaluations targeting attributes like fluency and relevance \cite{zheng2023judging,lin2023llm,fu2023gptscore,liu2023gpteval, kong2023large}. While these studies emphasize assessing existing dialogues, we also emphasize generating them, aiming for a closer alignment with human evaluations. Future work may merge LLM-based scoring for a more holistic evaluation approach.

\paragraph{LLM as Agents}
Previous research has explored the use of LLMs to simulate human behavior. These studies have employed LLMs for dataset construction, as seen in works such as \citep{wang2022self, xu2023baize, ding2023enhancing, li2023autoconv, kong2023platolm}
and to investigate interactions between agents \citep{park2023generative, li2023camel}.
Drawing inspiration from these endeavors, we explored the use of LLM agents in the dynamic evaluation of the human-machine conversation.

\section{Conclusion}
We have presented \ourmethod{} (\ourmed{}), a novel framework to evaluate AI assistants' API invocation capabilities through dynamic interactions. \ourmed{} utilizes a \useragent{} to emulate human patterns based on the provided \userscript{}.

Experiments on multiple assistants using our assembled benchmark showed \ourmed{} can reveal deficiencies overlooked by static evaluations. The \useragent{} aligned more closely with manual evaluation, uncovering issues like reluctance to invoke APIs and illusory parameter querying.

Overall, \ourmed{} serves as an effective automated substitute for expensive human assessment of assistants' API mastery. By mirroring human interactions, the user agent can dynamically prompt systems, identifying strengths and weaknesses invisible in static analyses.

\section{Ethics Statement}
The research presented in this paper focuses on developing a novel methodology for evaluating AI systems through simulated interactions. We have taken care to ensure our work is conducted ethically.

We have selected API domains that avoid potentially harmful or unethical use cases. The assembled benchmark comprises common assistant functionalities like device controls, media playback, scheduling, and information search. No APIs for surveillance, manipulation, or deception are included.

While this work aims to advance the state-of-the-art in AI evaluation, we recognize the potential for misuse. We advocate for development of robust benchmarks focused on beneficial applications, and for human oversight when deploying capable AI systems. With thoughtful implementation, improved evaluation techniques can guide progress towards trustworthy and helpful AI assistants.

\section{Limitations}

While promising, \ourmed{} has limitations to address. The benchmark dataset's small scale warrants expansion for generalizability. Incorporating qualitative metrics beyond accuracy, like coherence and naturalness, could provide a more holistic evaluation. Evaluating comprehension of API documentation itself could reveal insights into reasoning abilities. Lastly, results on closed-source commercial models hinders reproducibility.

\section{Acknowledgements}

We gratefully acknowledge the support of the National Natural Science Foundation of China (NSFC) via grant 62236004 and 62206078.

\section{Bibliographical References}\label{sec:reference}

\bibliographystyle{lrec-coling2024-natbib}
\bibliography{references}

\section{Language Resource References}
\label{lr:ref}
\bibliographystylelanguageresource{lrec-coling2024-natbib}
\bibliographylanguageresource{lr-references}

\end{document}